\documentclass{article}

\PassOptionsToPackage{numbers, compress}{natbib}
\usepackage[final]{nips_2017}

\usepackage[utf8]{inputenc} 
\usepackage[T1]{fontenc}    
\usepackage{hyperref}       
\usepackage{url}            
\usepackage{booktabs}       
\usepackage{amsfonts}       
\usepackage{nicefrac}       
\usepackage{microtype}      

\usepackage{amsmath}
\usepackage{graphicx}
\usepackage{caption}
\usepackage{subcaption}
\usepackage{multirow}
\usepackage[table]{xcolor}
\usepackage{bbm}
\def \Real{{\mathbb R}} 
 
\newcommand{\eValue}[1]{\mathbb{E}\left\{ #1 \right\}}

\newcommand{\cov}{\textrm{Cov}}

\newcommand{\ie}{i.e., }

\newcommand{\eg}{e.g. }

\newcommand{\N}{\mathcal{N}}

\title{Towards Personalized Modeling of the Female Hormonal Cycle: Experiments with Mechanistic Models and Gaussian Processes}

\author{
        I\~{n}igo Urteaga \\
        Columbia University; New York City, NY\\
        \texttt{inigo.urteaga@columbia.edu} \\
        \And
        David J. Albers \\
        Columbia University; New York City, NY \\
        \texttt{dja2119@columbia.edu} \\
        \And    
        Marija Vlajic Wheeler \\
        Clue; Berlin, Germany \\
        \texttt{marija@helloclue.com}\\
        \And
        Anna Druet \\
        Clue; Berlin, Germany \\
        \texttt{anna@helloclue.com} \\
        \And
        Hans Raffauf \\
        Clue; Berlin, Germany \\
        \texttt{hans@helloclue.com} \\
        \And
        No\'{e}mie Elhadad \\
        Columbia University; New York City, NY \\
        \texttt{noemie.elhadad@columbia.edu}
}

\begin{document}

\maketitle

\begin{abstract}
In this paper, we introduce a novel task for machine learning in healthcare, namely personalized modeling of the female hormonal cycle. The motivation for this work is to model the hormonal cycle and predict its phases in time, both for healthy individuals and for those with disorders of the reproductive system. Because there are individual differences in the menstrual cycle, we are particularly interested in personalized models that can account for individual idiosyncracies, towards identifying phenotypes of menstrual cycles. As a first step, we consider the hormonal cycle as a set of observations through time. We use a previously validated mechanistic model to generate realistic hormonal patterns, and experiment with Gaussian process regression to estimate their values over time. Specifically, we are interested in the feasibility of predicting menstrual cycle phases under varying learning conditions: number of cycles used for training, hormonal measurement noise and sampling rates, and informed vs. agnostic sampling of hormonal measurements. Our results indicate that Gaussian processes can help model the female menstrual cycle. We discuss the implications of our experiments in the context of modeling the female menstrual cycle. 
\end{abstract}

\section{Introduction}
\label{sec:intro}

We are interested in modeling the hormones regulating the female menstrual cycle using machine learning. Modeling the female reproductive endocrine system can be useful in several ways. First, knowledge about reproductive hormones directly informs and predicts phases of the menstrual cycle; such predictions are useful for women in a variety of contexts, including conception, contraception and identifying abnormal phase durations \citep{j-Jordan1994, j-Crawford2017}. Second, there are several chronic, hormone-dependent disorders of the female reproductive system, including polycystic ovarian syndrome and endometriosis, two particularly prevalent disorders \citep{j-Carmina1999,j-Giudice2010,j-Barbosa2011}. Modeling reproductive hormones can yield valuable insight into these conditions by identifying hormonal signatures across individuals that correspond to phenotypes of disease. Further, hormone modeling could contribute to diagnosing and even predicting which individuals have or are at risk of having a disorder. Third, a personalized model of the reproductive endocrine system can help predicting individual responses to drug exposures (e.g., birth control and other hormonal therapies) and environmental exposures (e.g., endocrine disruptors). Finally, while the reproductive endocrine system has been actively investigated in biology and physiology, there is no existing model that characterizes the reproductive hormones in the context of other phenotypic, easily observable or measurable variables (e.g., signs and symptoms such as pelvic pain during periods, moods variation, body weight fluctuations, or period length). A comprehensive model of the reproductive endocrine cycle that relates physiologic and phenotypic variables has the potential to yield robust, personalized models that can infer hormones dynamics and cycle phases through minimally invasive measurements.

Menstrual trackers provide a particularly exciting opportunity for collecting
phenotypic realizations of the menstrual cycle. For instance, through the
popular menstrual tracker Clue, we have access to a large population of women
with a natural cycle (i.e., not exposed to any hormonal treatment), and who
track their cycle along with a variety of phenotypic variables, including
``symptoms'' like cramps, moods, and behaviors like exercise and sleep. In this
dataset, about 750,000 women have tracked at least 4 cycles. Furthermore, about
70,000 women tracked at least one positive ovulation test. In aggregate, these
data can help infer population-level models for phenotypic variables. When
coupled with hormone measurements, they can help train robust machine learning
models that jointly characterize the interactions between hormones and
phenotypic variables. In this paper, we focus on modeling the dynamics of the
hormonal cycle, and postpone its connection to other phenotypic variables for
future work. 

The female reproductive endocrine system is complex, operates over multiple
time scales, and varies amongst individuals. Researchers have studied personal
variability of the menstrual cycle \citep{j-Lenton1984,j-Alliende2002},
populations and subpopulations of  women with regular periods
\citep{j-Landgren1980,j-Bonen1981,j-Ecochard2017}, the relationship between
hormonal phases and both fertility \citep{j-Jordan1994, j-Crawford2017} and menopause
\citep{j-Prior1998,j-Landgren2004,j-Prior2011}, and how behaviors like diet
\citep{j-Barr1995} and exercise \citep{j-Prior1987} can impact hormonal cycles.
Nevertheless, the complexity of the hormonal dynamics and the invasive nature
of their measurement make a more complete understanding of the female menstrual
cycle difficult. 

The female reproductive hormonal cycle has been the focus of mathematical systems physiology \cite{b-Keener2009}, and several mechanistic models have been proposed in the literature. These models are systems of non-linear delay differential equations that describe how hormones interact with each other and through time. The state-of-the-art model proposed by \citet{j-Clark2003} has been shown to fit well when tried against a dataset of daily hormonal measurements of healthy individuals with natural, regular cycles. That is, the mathematical model captures the cycle physiology with high fidelity.

The literature also contains a wide spectrum of methodologies for collecting and investigating the hormonal cycle, ranging from large-scale cohort analysis with ``shallow'' hormone measurements (under-sampling in time, or sampling only at known phases of the cycle with proxies to blood tests), to small-scale cohort analysis with ``in-depth'' hormone measurements (daily serum readings throughout the menstrual cycle). There is an opportunity for machine learning models to determine the data requirements, and in particular the hormone sampling rate, that can mitigate the tension between the need for multiple measurements and invasive measurements. 

In this paper, we focus on a set of feasibility experiments, as a first step towards modeling of the reproductive cycle through machine learning. We use established mathematical models to generate \emph{in-silico} data, and use a Gaussian process model (GPM) \cite{b-Rasmussen2006} to answer the following questions: (i) is it possible to reconstruct and
predict the hormonal values over time, and the associated menstrual cycle phases using a GPM?; (ii) how sensitive is the  GPM to sampling frequency?; and (iii) how sensitive is the GPM to measurement noise?

The main contributions of this work are as follows: (1) we tackle a new task, namely studying female reproductive cycle through learning from hormonal measurements; (2) we introduce the use of mechanistic models from the fields of mathematical physiology and biology to generate realistic training datasets that truthfully capture the menstrual cycle physiology; (3) we explore the data requirements of a Gaussian process to accurately forecast hormonal values and identify the phases within the menstrual cycle.

\section{Mechanistic model and in-silico data for the hormonal menstrual cycle}
\label{sec:hmc_model}

A normal menstrual cycle for an adult woman consists of two phases, the follicular and the luteal phases, separated by ovulation. The pituitary synthesizes and releases the follicle stimulating hormone ($FSH$) and the luteinizing hormone ($LH$), impacting the blood levels of $LH$ and $FSH$ to which the ovary responds. The ovary produces estradiol ($E_2$), progesterone ($P_4$), and inhibin ($I_h$), which control the pituitary’s synthesis and release of the gonadotropin hormones during the various stages of the cycle. Here we use a model developed by \citet{j-Clark2003}, a 13-dimensional delayed differential equation shown in Appendix \ref{app:mech_model} that merges the pituitary \citep{j-Schlosser2000} and ovarian \citep{j-Selgrade1999} hormonal models for daily average hormone concentrations. $LH$ and $FSH$ denote the serum concentrations of each hormone, and $LH_{RP}$ and $FSH_{RP}$ represent their concentration in the reserve pools. The variables $LH$ and $FSH$ depend nonlinearly on the ovarian states and have delay dependencies that control the synthesizing and inhibiting effects of $E_2$, $P_4$ and $I_h$. Full details of the physiological explanation for the model can be found in \citep{j-Clark2003,j-Selgrade2009}, where (1) its consistency and sensitivity with regard to real datasets has been validated, and (2) its asymptotically stable periodic solutions for both healthy and unhealthy patterns of female hormonal cycles are shown to exist (\ie it closely captures the physiology of real menstrual cycle datasets). 

\begin{figure}[!h]
        \centering
        \begin{subfigure}[b]{0.41\textwidth}
                \includegraphics[width=\textwidth]{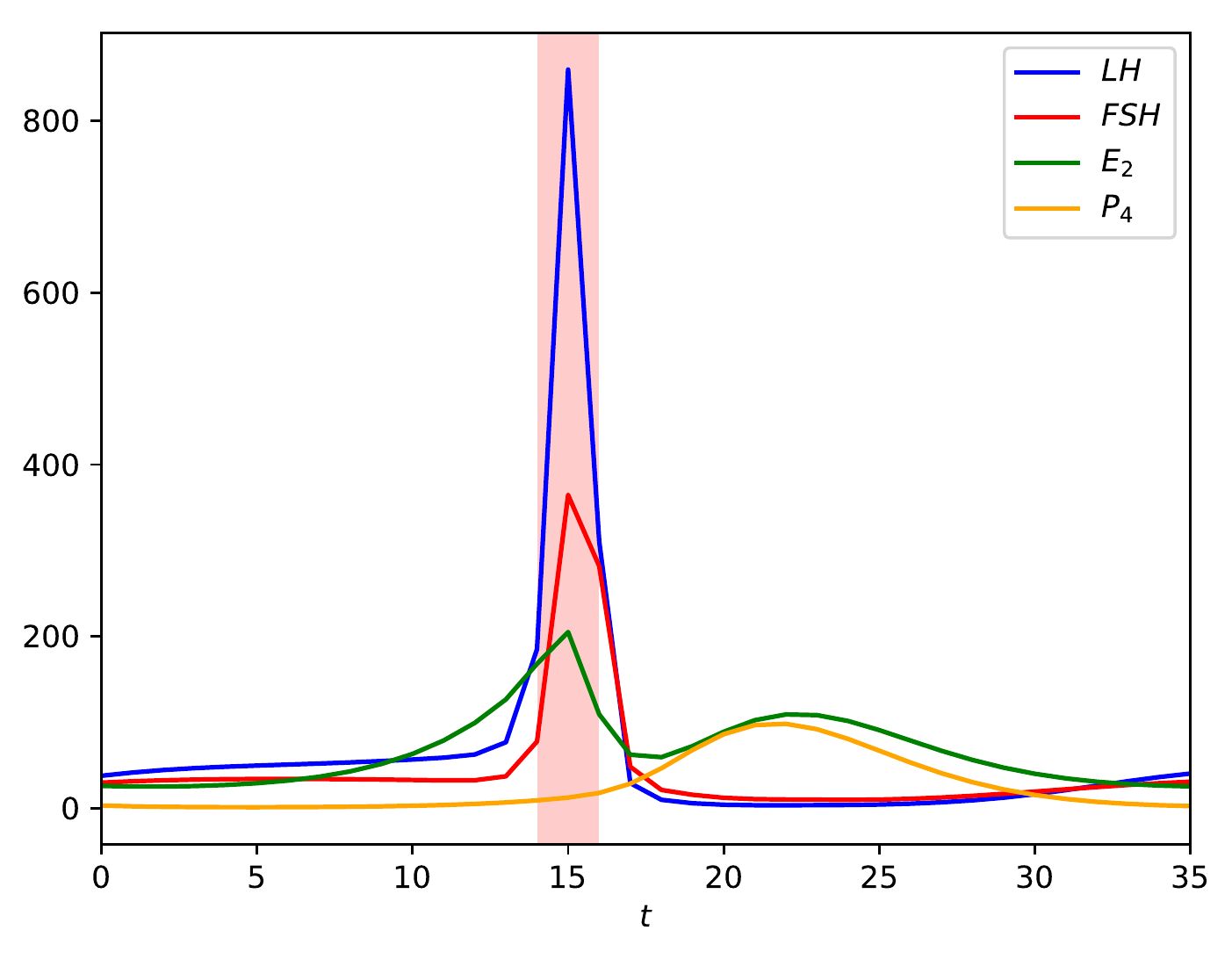}
                \vspace*{-0.5cm}
                \caption{Evolution over time of the hormones of interest. Ovulation (in red) separates the follicular and luteal phases.}               
                \label{fig:hmc_data_example}
        \end{subfigure}
        \qquad
        \begin{subfigure}[b]{0.4\textwidth}
                \includegraphics[width=\textwidth]{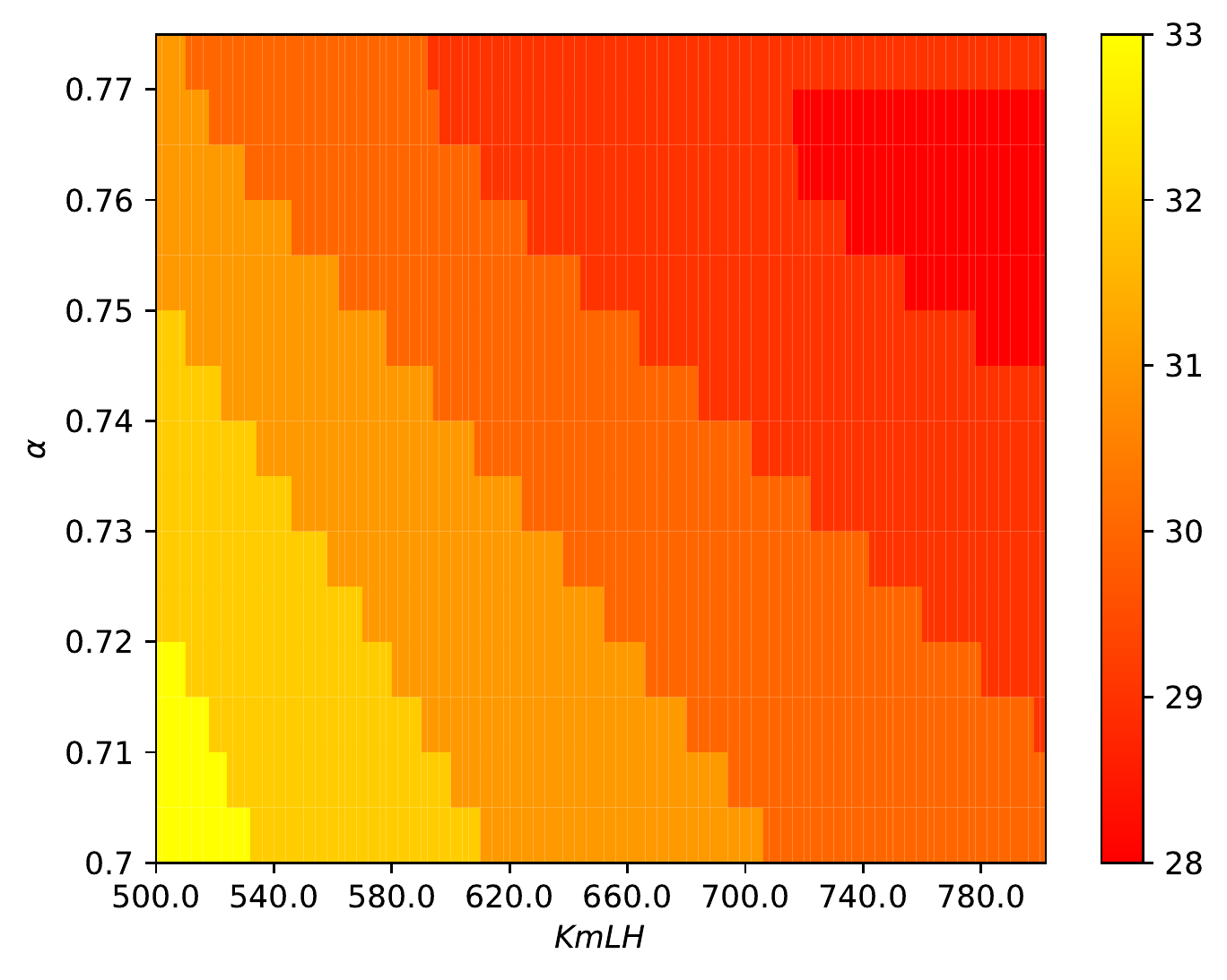}
                \vspace*{-0.5cm}
                \caption{Periodicity, in days, of the menstrual cycle data as a function of the $LH$ synthesis parameters $\alpha$ and $Km_{LH}$.}
                \label{fig:hmc_data_periods}
        \end{subfigure}
        \caption{Simulated hormonal menstrual cycle data.}
        \label{fig:hmc_data}
\end{figure}

In-silico data are generated from the set of equations provided in Appendix \ref{app:mech_model} with parameters as described in \citet{j-Clark2003}, using the delayed-differential equation solver developed by \citet{j-Shampine2001}. Figure \ref{fig:hmc_data_example} shows a typical menstrual cycle as generated by the mechanistic model, with both the hormone evolution over time and the follicular, ovulation and luteal phases. To obtain personalized menstrual patterns, we vary the parameter set controlling the model, thus matching the physiology of diverse women cycles. In particular, we focus on the $LH$ synthesis parameters (following \citet{j-Selgrade2009}) to generate menstrual cycles with slightly different characteristics, as shown in Fig. \ref{fig:hmc_data_periods}.

\vspace*{-0.5cm}
\section{Gaussian process based modeling of the  menstrual cycle}
\label{sec:hmc_gpm}

\paragraph*{Gaussian Processes.} Gaussian processes provide a Bayesian nonparametric approach to smoothing and forecasting by imposing a prior directly over functions, rather than parameters \citep{ic-MacKay1998}. The distribution over functions is defined as $f(x) \sim GP\left(m(x), k(x,x')\right)$, where $x\in \Real^d$ is an arbitrary input variable. A GPM is fully determined by its mean $m(x)=\eValue{f(x)}$ and covariance (or kernel) $k(x,x')=\cov\left\{f(x), f(x')\right\}$ functions. In practice, one does not directly observe the function values $f(x_1), f(x_2),\cdots,f(x_N)$, but their noisy version $y_{1:N} \equiv \left\{y_1, y_2,\cdots,y_N\right\}$, where $y_n=f(x_n)+\epsilon_n$ and $\epsilon \sim p(\epsilon)$. If the observation noise is Gaussian, \ie $p(\epsilon)=\N\left(\epsilon |\mu_{\epsilon}, \sigma_{\epsilon}^2\right)$, one can exactly infer the posterior predictive distribution over the underlying Gaussian process, and the marginal likelihood of the observed data. That is, one computes $p(y_{1:N}|x_{1:N}, \theta)$, where $\theta$ refers to the
hyperparameters of the Gaussian process, in closed form \citep{b-Cressie1993, b-Stein1999, b-Rasmussen2006}. The properties of the functions induced by a Gaussian process (\ie smoothness, periodicity, etc.) are fully determined by the kernel function $k(x,x')$, and the accuracy of the GPM is dependent on the appropriate choice of kernel \citep{ic-MacKay1998,tr-Abrahamsen1997,b-Rasmussen2006}. Because we are modeling a system whose data are periodic and smooth, we use a combination of rational quadratic and periodic kernels (see Appendix \ref{app:gpm_kernels} for details).

\paragraph*{Experimental Setup.} The model in Appendix \ref{app:mech_model} is used to simulate \textit{ground truth} hormonal cycle patterns. We mimic realistic hormonal measurements by downgrading the true values both in quality (additive white noise) and availability (downsampling). We train a per-hormone GPM with time as input (via \citet{j-Neumann2015}), for particular realizations of noise, sampling rates, and training intervals. To identify the phases within the menstrual cycle, we process the output of the GPM to find hormonal minima and maxima with a simple above (below) the mean plus (minus) one standard deviation heuristic. Our evaluation metric is prediction accuracy, \ie the ratio of predicted days to ground truth days for a given phase.

\begin{figure}[!h]
        \centering
        \begin{subfigure}[b]{0.42\textwidth}
                \includegraphics[width=\textwidth]{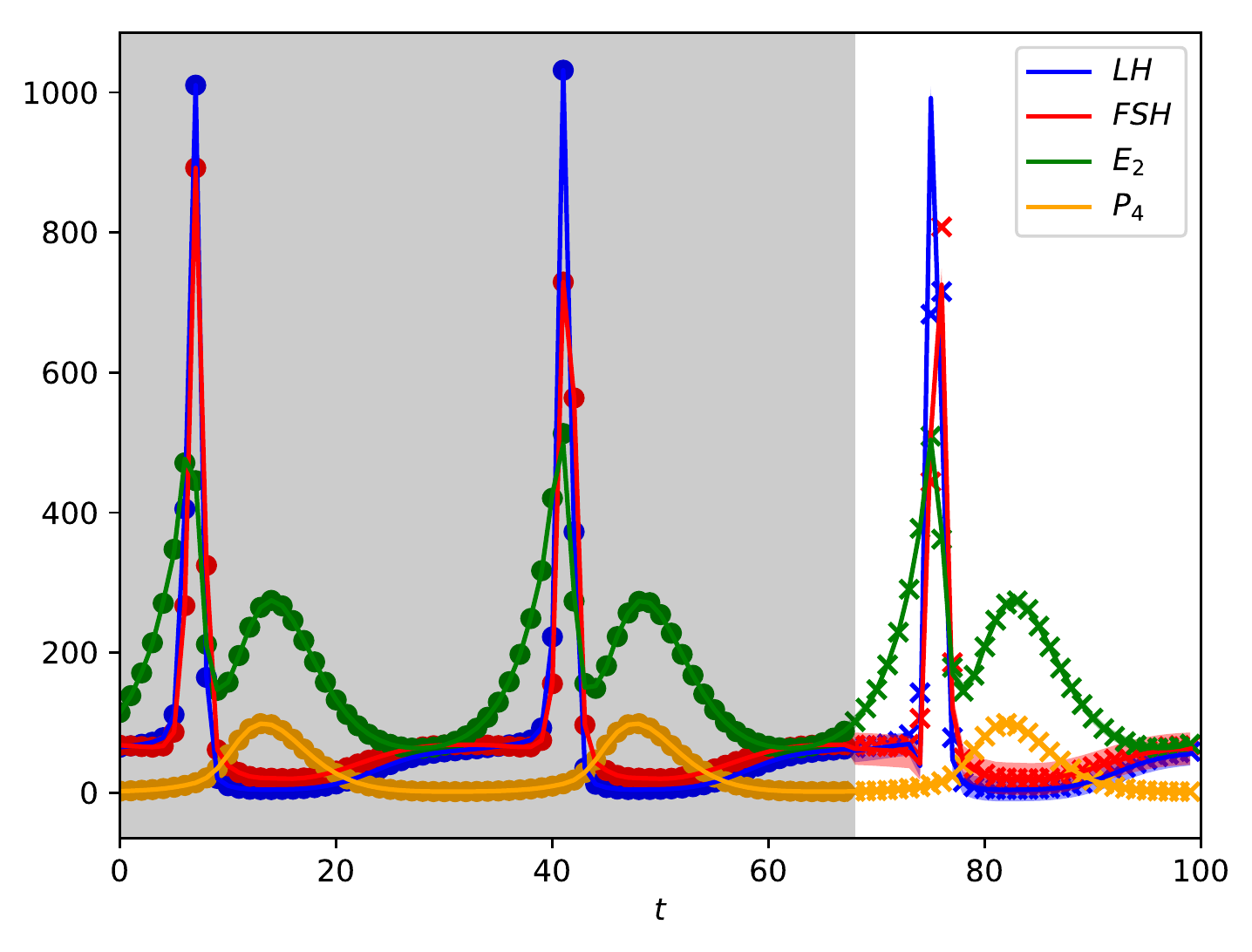}
                \vspace*{-0.7cm}
                \caption{Daily training data.}
                \label{fig:hmc_predicted_gp_rate_1}
        \end{subfigure}
        \begin{subfigure}[b]{0.42\textwidth}
                \includegraphics[width=\textwidth]{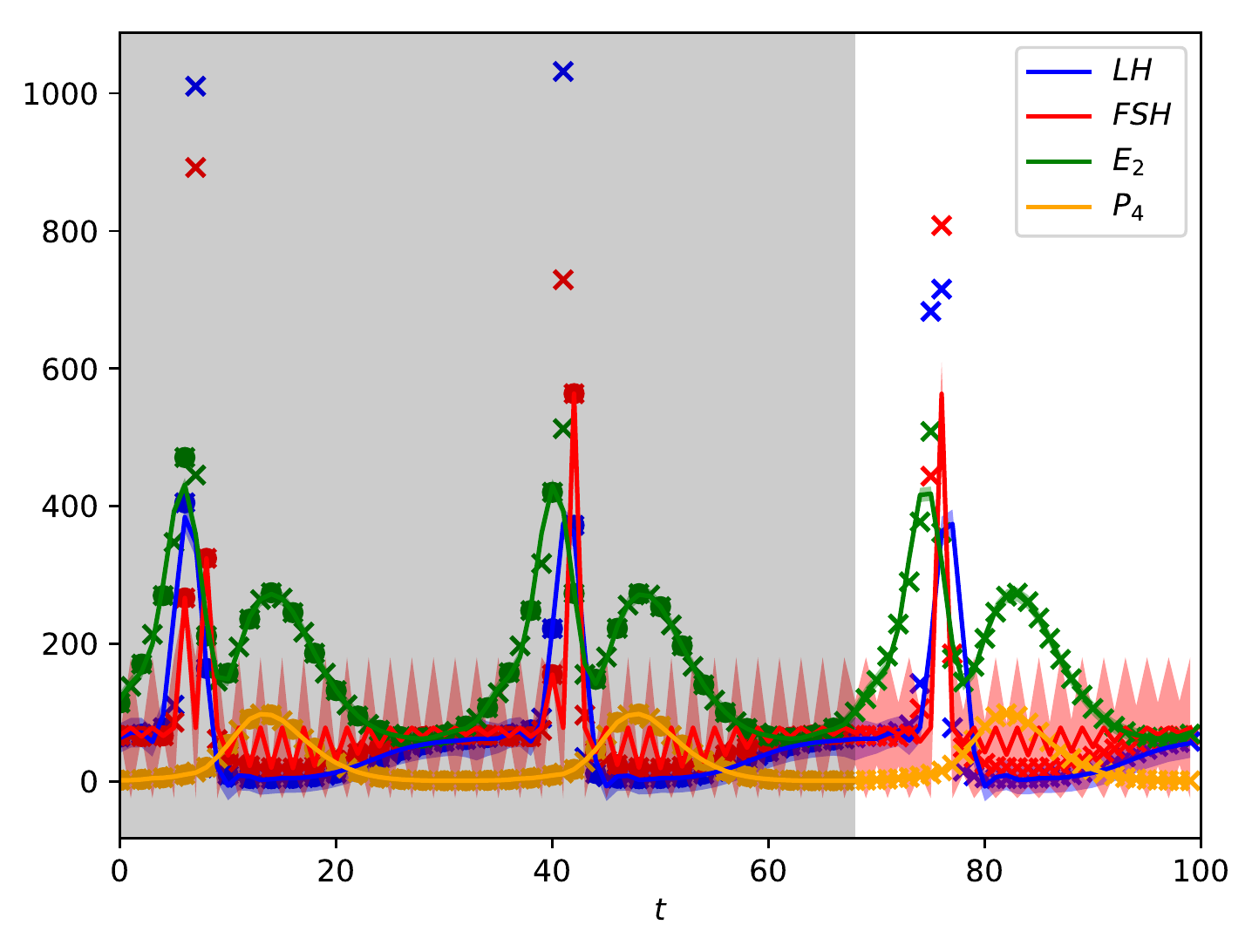}
                \vspace*{-0.7cm}
                \caption{Once every two days training data.}
                \label{fig:hmc_predicted_gp_rate_2}
        \end{subfigure}
        \vspace*{-0.2cm}
		\caption{Ground truth in-silico data (crosses), training data points (dots) and GP estimated values (solid lines). The training interval is shown as the shaded region.}
        \label{fig:hmc_predicted_gp}
\end{figure}

\paragraph*{Results.} Fig. \ref{fig:hmc_predicted_gp} shows that the GPM makes accurate \emph{personalized} predictions when observing all data points without noise, and underestimates extreme hormonal values when peaks are not observed in the training set. This predictive performance is only possible if at least \emph{two} peaks of the hormonal cycles are included in the training set. Tables \ref{tab:event_prediction_rate} and \ref{tab:event_prediction_noise} in Appendix \ref{app:event_prediction_tables} provide a complete set of summary statistics of the GPM's accuracy on predicting menstrual cycle phases.

Table \ref{tab:event_prediction_rate} illustrates the impact of downsampling, and shows a deterioration when less than two samples per week are measured, even without noise. Table \ref{tab:event_prediction_noise} shows the impact of noise and how GPMs are robust to uncertainty. When the noisy measurements obscure the hormonal peaks, predicting the events of interest becomes difficult. The critical factor for GPM's performance is the presence of hormonal peak measurements in the training set (\eg see $LH$ peak prediction errors in Fig. \ref{fig:hmc_predicted_gp_rate_2}). The performance is improved when at least one sample from the peak is included in the training interval (\ie informed sampling), and suggests that sparse knowledge-targeted measurements may provide useful. High-accuracy predictive GPMs are promising for diagnostic and discriminative purposes (\ie identifying normal/abnormal phenotypes).

\vspace*{-0.25cm}
\section{Discussion}
\vspace*{-0.5em}
We have explored the feasibility of GPMs to forecast personalized female
hormonal values. GPMs are an established, robust, and simple method with
successful applications in the biomedical domain
\citep{ip-Duerichen2014,ip-Pimentel2013}, and in modeling physiological
time-series in particular \citep{j-Duerichen2015,j-Cheng2017}. Presented
results suggest that our machine-learning approach is accurate, and allows for
identification of the phases within the menstrual cycle. Knowledge-based
adaptive sampling produces fewer but more informative measurements for GPM
prediction, indicative that fewer invasive hormone measurements can be required
in practice. In our future work, we will first leverage the dependencies
between hormones via multi-output Gaussian processes
\cite{ic-Boyle2005,ic-Alvarez2009,ic-Bonilla2008}. Second, we will investigate
the use of additional variables in modeling the hormonal cycle, including
phenotypic variables like the ones typically tracked in menstrual trackers. We
hypothesize that these will bring more comprehensive models, all the while
putting forth the connection between hormones and signs, behaviors, and
symptoms through time. We envision that these models will help understand the
female menstrual cycle and identify phenotypes, both for disease diagnoses and
discrimination amongst individuals. Continued research on machine learning
techniques will have significant impact on the study of women's hormone
health, and we are eager for the broader machine learning community to adopt
this problem.\footnote{Code for the mechanistic model, the Gaussian processes,
  and the dataset of simulated menstrual cycles is available at \url{https://github.com/iurteaga/hmc}.} 

%

\bibliographystyle{plainnat}
\bibliography{../references}

\begin{thebibliography}{34}
\providecommand{\natexlab}[1]{#1}
\providecommand{\url}[1]{\texttt{#1}}
\expandafter\ifx\csname urlstyle\endcsname\relax
  \providecommand{\doi}[1]{doi: #1}\else
  \providecommand{\doi}{doi: \begingroup \urlstyle{rm}\Url}\fi

\bibitem[Abrahamsen(1997)]{tr-Abrahamsen1997}
Petter Abrahamsen.
\newblock {A review of Gaussian random fields and correlation functions}.
\newblock Technical report, Norwegian Computing Center, 1997.
\newblock URL \url{http://publications.nr.no/917_Rapport.pdf}.

\bibitem[Alliende(2002)]{j-Alliende2002}
Mar\'{i}a~Elena Alliende.
\newblock {Mean versus individual hormonal profiles in the menstrual cycle}.
\newblock \emph{Fertility and Sterility}, 78\penalty0 (1):\penalty0 90 -- 95,
  2002.
\newblock URL
  \url{http://www.sciencedirect.com/science/article/pii/S0015028202031679}.

\bibitem[Alvarez and Lawrence(2009)]{ic-Alvarez2009}
Mauricio Alvarez and Neil~D. Lawrence.
\newblock {Sparse Convolved Gaussian Processes for Multi-output Regression}.
\newblock In D.~Koller, D.~Schuurmans, Y.~Bengio, and L.~Bottou, editors,
  \emph{Advances in Neural Information Processing Systems 21}, pages 57--64.
  Curran Associates, Inc., 2009.
\newblock URL
  \url{http://papers.nips.cc/paper/3553-sparse-convolved-gaussian-processes-for-multi-output-regression.pdf}.

\bibitem[Barbosa et~al.(2011)Barbosa, Souza, Bianco, and
  Christofolini]{j-Barbosa2011}
C~Parente Barbosa, AM~Bentes~De Souza, B~Bianco, and DM~Christofolini.
\newblock {The effect of hormones on endometriosis development}.
\newblock \emph{Minerva Ginecologica: A Journal on Obstetrics and Gynecology},
  63\penalty0 (4):\penalty0 375 -- 386, Aug 2011.
\newblock URL
  \url{https://www.minervamedica.it/en/journals/minerva-ginecologica/article.php?cod=R09Y2011N04A0375}.

\bibitem[Barr et~al.(1995)Barr, Janelle, and Prior]{j-Barr1995}
S.I. Barr, K.C. Janelle, and Jerilynn~C. Prior.
\newblock {Energy intakes are higher during the luteal phase of ovulatory
  menstrual cycles}.
\newblock \emph{The American Journal of Clinical Nutrition}, 61\penalty0
  (1):\penalty0 39--43, 1995.
\newblock URL \url{http://ajcn.nutrition.org/content/61/1/39.abstract}.

\bibitem[Bonen et~al.(1981)Bonen, Belcastro, Ling, and Simpson]{j-Bonen1981}
A.~Bonen, A.~N. Belcastro, W.~Y. Ling, and A.~A. Simpson.
\newblock {Profiles of selected hormones during menstrual cycles of teenage
  athletes}.
\newblock \emph{Journal of Applied Physiology}, 50\penalty0 (3):\penalty0
  545--551, 1981.
\newblock URL \url{http://jap.physiology.org/content/50/3/545}.

\bibitem[Bonilla et~al.(2008)Bonilla, Chai, and Williams]{ic-Bonilla2008}
Edwin~V Bonilla, Kian~M. Chai, and Christopher Williams.
\newblock {Multi-task Gaussian Process Prediction}.
\newblock In J.~C. Platt, D.~Koller, Y.~Singer, and S.~T. Roweis, editors,
  \emph{Advances in Neural Information Processing Systems 20}, pages 153--160.
  Curran Associates, Inc., 2008.
\newblock URL
  \url{http://papers.nips.cc/paper/3189-multi-task-gaussian-process-prediction.pdf}.

\bibitem[Boyle and Frean(2005)]{ic-Boyle2005}
Phillip Boyle and Marcus Frean.
\newblock {Dependent Gaussian Processes}.
\newblock In L.~K. Saul, Y.~Weiss, and L.~Bottou, editors, \emph{Advances in
  Neural Information Processing Systems 17}, pages 217--224. MIT Press, 2005.
\newblock URL
  \url{http://papers.nips.cc/paper/2561-dependent-gaussian-processes.pdf}.

\bibitem[Carmina and Lobo(1999)]{j-Carmina1999}
Enrico Carmina and Rogerio~A. Lobo.
\newblock {Polycystic Ovary Syndrome (PCOS): Arguably the Most Common
  Endocrinopathy Is Associated with Significant Morbidity in Women}.
\newblock \emph{The Journal of Clinical Endocrinology \& Metabolism},
  84\penalty0 (6):\penalty0 1897--1899, 1999.
\newblock URL \url{http://dx.doi.org/10.1210/jcem.84.6.5803}.

\bibitem[Cheng et~al.(2017)Cheng, Darnell, Chivers, Draugelis, Li, and
  Engelhardt]{j-Cheng2017}
Li-Fang Cheng, Gregory Darnell, Corey Chivers, Michael~E Draugelis, Kai Li, and
  Barbara~E Engelhardt.
\newblock {Sparse Multi-Output Gaussian Processes for Medical Time Series
  Prediction}.
\newblock \emph{ArXiv e-prints}, 2017.

\bibitem[Clark et~al.(2003)Clark, Schlosser, and Selgrade]{j-Clark2003}
Leona~Harris Clark, Paul~M. Schlosser, and James~F. Selgrade.
\newblock {Multiple stable periodic solutions in a model for hormonal control
  of the menstrual cycle}.
\newblock \emph{Bulletin of Mathematical Biology}, 65\penalty0 (1):\penalty0
  157--173, Jan 2003.
\newblock URL \url{https://doi.org/10.1006/bulm.2002.0326}.

\bibitem[Crawford et~al.(2017)Crawford, Pritchard, Herring, and
  Steiner]{j-Crawford2017}
Natalie~M. Crawford, David~A. Pritchard, Amy~H. Herring, and Anne~Z. Steiner.
\newblock {Prospective evaluation of luteal phase length and natural
  fertility}.
\newblock \emph{Fertility and Sterility}, 107\penalty0 (3):\penalty0 749 --
  755, 2017.
\newblock URL
  \url{http://www.sciencedirect.com/science/article/pii/S0015028216630224}.

\bibitem[Cressie(1993)]{b-Cressie1993}
Noel~A.C. Cressie.
\newblock \emph{{Statistics for spatial data}}.
\newblock Wiley series in probability and mathematical statistics: Applied
  probability and statistics. J. Wiley, 1993.
\newblock ISBN 9780471002550.

\bibitem[D\"{u}richen et~al.(2014)D\"{u}richen, Pimentel, Clifton, Schweikard,
  and Clifton]{ip-Duerichen2014}
R.~D\"{u}richen, M.~A.~F. Pimentel, L.~Clifton, A.~Schweikard, and D.~A.
  Clifton.
\newblock {Multi-task Gaussian process models for biomedical applications}.
\newblock In \emph{IEEE-EMBS International Conference on Biomedical and Health
  Informatics (BHI)}, pages 492--495, June 2014.

\bibitem[D\"{u}richen et~al.(2015)D\"{u}richen, Pimentel, Clifton, Schweikard,
  and Clifton]{j-Duerichen2015}
R.~D\"{u}richen, M.~A.~F. Pimentel, L.~Clifton, A.~Schweikard, and D.~A.
  Clifton.
\newblock {Multitask Gaussian Processes for Multivariate Physiological
  Time-Series Analysis}.
\newblock \emph{IEEE Transactions on Biomedical Engineering}, 62\penalty0
  (1):\penalty0 314--322, Jan 2015.

\bibitem[Ecochard et~al.(2017)Ecochard, Bouchard, Leiva, Abdulla, Dupuis,
  Duterque, Billard, Boehringer, and Genolini]{j-Ecochard2017}
Rene Ecochard, Thomas Bouchard, Rene Leiva, Saman Abdulla, Olivier Dupuis,
  Olivia Duterque, Marie~Garmier Billard, Hans Boehringer, and Christophe
  Genolini.
\newblock {Characterization of hormonal profiles during the luteal phase in
  regularly menstruating women}.
\newblock \emph{Fertility and Sterility}, 108\penalty0 (1):\penalty0 175 --
  182.e1, 2017.
\newblock URL
  \url{http://www.sciencedirect.com/science/article/pii/S0015028217303692}.

\bibitem[Giudice(2010)]{j-Giudice2010}
Linda~C. Giudice.
\newblock {Endometriosis}.
\newblock \emph{New England Journal of Medicine}, 362\penalty0 (25):\penalty0
  2389 -- 2398, 2010.
\newblock URL \url{http://dx.doi.org/10.1056/NEJMcp1000274}.

\bibitem[Jordan et~al.(1994)Jordan, Craig, Clifton, and Soules]{j-Jordan1994}
John Jordan, Kristin Craig, Donald~K. Clifton, and Michael~R. Soules.
\newblock {Luteal phase defect: the sensitivity and specificity of diagnostic
  methods in common clinical use}.
\newblock \emph{Fertility and Sterility}, 62\penalty0 (1):\penalty0 54 -- 62,
  1994.
\newblock URL
  \url{http://www.sciencedirect.com/science/article/pii/S0015028216568150}.

\bibitem[Keener and Sneyd(2009)]{b-Keener2009}
James Keener and Alfred James~Robert Sneyd.
\newblock \emph{{Mathematical Physiology II: Systems Physiology}}.
\newblock Springer-Verlag New York, 2009.

\bibitem[Landgren et~al.(1980)Landgren, Und\'{e}n, and
  Diczfalusy]{j-Landgren1980}
B.-M. Landgren, A.-L. Und\'{e}n, and E.~Diczfalusy.
\newblock {Hormonal profile of the cycle in 68 normally menstruating women}.
\newblock \emph{Acta Endocrinologica}, 94\penalty0 (1):\penalty0 89--98, 1980.
\newblock URL \url{http://www.eje-online.org/content/94/1/89.abstract}.

\bibitem[Landgren et~al.(2004)Landgren, Collins, Csemiczky, Burger, Baksheev,
  and Robertson]{j-Landgren2004}
Britt-Marie Landgren, Aila Collins, Giorgy Csemiczky, Henry~G. Burger, Lyrissa
  Baksheev, and David~M. Robertson.
\newblock {Menopause Transition: Annual Changes in Serum Hormonal Patterns over
  the Menstrual Cycle in Women during a Nine-Year Period Prior to Menopause}.
\newblock \emph{The Journal of Clinical Endocrinology \& Metabolism},
  89\penalty0 (6):\penalty0 2763--2769, 2004.
\newblock URL \url{+ http://dx.doi.org/10.1210/jc.2003-030824}.

\bibitem[Lenton et~al.(1984)Lenton, Landgren, and Sexton]{j-Lenton1984}
Elizabeth~A. Lenton, Brut-Marie Landgren, and Lynne Sexton.
\newblock {Normal variation in the length of the luteal phase of the menstrual
  cycle: identification of the short luteal phase}.
\newblock \emph{BJOG: An International Journal of Obstetrics \& Gynaecology},
  91\penalty0 (7):\penalty0 685--689, 1984.
\newblock URL \url{http://dx.doi.org/10.1111/j.1471-0528.1984.tb04831.x}.

\bibitem[MacKay and David(1998)]{ic-MacKay1998}
David MacKay and J.C. David.
\newblock {Introduction to Gaussian processes}.
\newblock In Christopher~M. Bishop, editor, \emph{{Neural Networks and Machine
  Learning}}, volume 168, chapter~11, pages 133 -- 165. Springer-Verlag, 1998.
\newblock URL \url{http://www.inference.org.uk/mackay/gpB.pdf}.

\bibitem[Neumann et~al.(2015)Neumann, Huang, Marthaler, and
  Kersting]{j-Neumann2015}
Marion Neumann, Shan Huang, Daniel~E. Marthaler, and Kristian Kersting.
\newblock {pyGPs -- A Python Library for Gaussian Process Regression and
  Classification}.
\newblock \emph{Journal of Machine Learning Research}, 16:\penalty0 2611--2616,
  2015.
\newblock URL \url{http://jmlr.org/papers/v16/neumann15a.html}.

\bibitem[Pimentel et~al.(2013)Pimentel, Clifton, Clifton, and
  Tarassenko]{ip-Pimentel2013}
Marco~A.F. Pimentel, David~A. Clifton, Lei Clifton, and Lionel Tarassenko.
\newblock {Modelling Patient Time-Series Data from Electronic Health Records
  using Gaussian Processes}.
\newblock In \emph{NIPS Workshop on Machine Learning for Clinical Data
  Analysis}, 2013.
\newblock URL \url{http://www.robots.ox.ac.uk/~davidc/pubs/nips2013.pdf}.

\bibitem[Prior(1998)]{j-Prior1998}
Jerilynn~C. Prior.
\newblock {Perimenopause: The Complex Endocrinology of the Menopausal
  Transition}.
\newblock \emph{Endocrine Reviews}, 19\penalty0 (4):\penalty0 397--428, 1998.
\newblock URL \url{+ http://dx.doi.org/10.1210/edrv.19.4.0341}.

\bibitem[Prior and Hitchcock(2011)]{j-Prior2011}
Jerilynn~C. Prior and Christine~L. Hitchcock.
\newblock {The endocrinology of perimenopause: need for a paradigm shift}.
\newblock \emph{Frontiers in bioscience (Scholar edition)}, 3:\penalty0 474 --
  486, 2011.
\newblock URL \url{https://doi.org/10.2741/s166}.

\bibitem[Prior et~al.(1987)Prior, Vigna, Sciarretta, Alojado, and
  Schulzer]{j-Prior1987}
Jerilynn~C. Prior, Yvette Vigna, Danielle Sciarretta, Nenita Alojado, and
  Michael Schulzer.
\newblock {Conditioning exercise decreases premenstrual symptoms: a
  prospective, controlled 6-month trial}.
\newblock \emph{Fertility and Sterility}, 47\penalty0 (3):\penalty0 402 -- 408,
  1987.
\newblock URL
  \url{http://www.sciencedirect.com/science/article/pii/S0015028216590451}.

\bibitem[Rasmussen and Williams(2006)]{b-Rasmussen2006}
Carl~Edward Rasmussen and Christopher K.~I. Williams.
\newblock \emph{{Gaussian Processes for Machine Learning}}.
\newblock The MIT Press, 2006.
\newblock URL \url{http://www.gaussianprocess.org/gpml/}.

\bibitem[Schlosser and Selgrade(2000)]{j-Schlosser2000}
Paul~M. Schlosser and James~F. Selgrade.
\newblock {A model of gonadotropin regulation during the menstrual cycle in
  women: qualitative features}.
\newblock \emph{Environmental Health Perspectives}, October 2000.
\newblock URL \url{https://www.ncbi.nlm.nih.gov/pubmed/11035997}.

\bibitem[Selgrade and Schlosser(1999)]{j-Selgrade1999}
James~F. Selgrade and Paul~M. Schlosser.
\newblock {A model for the production of ovarian hormones during the menstrual
  cycle}.
\newblock \emph{Fields Institute Communications}, pages 429 -- 446, 1999.
\newblock URL
  \url{http://citeseerx.ist.psu.edu/viewdoc/download?doi=10.1.1.149.7415&rep=rep1&type=pdf}.

\bibitem[Selgrade et~al.(2009)Selgrade, Harris, and Pasteur]{j-Selgrade2009}
James~F. Selgrade, L.A. Harris, and R.D. Pasteur.
\newblock {A model for hormonal control of the menstrual cycle: Structural
  consistency but sensitivity with regard to data}.
\newblock \emph{Journal of Theoretical Biology}, 260\penalty0 (4):\penalty0 572
  -- 580, 2009.
\newblock URL
  \url{http://www.sciencedirect.com/science/article/pii/S0022519309002835}.

\bibitem[Shampine and Thompson(2001)]{j-Shampine2001}
L.F. Shampine and S.~Thompson.
\newblock {Solving DDEs in Matlab}.
\newblock \emph{Applied Numerical Mathematics}, 37\penalty0 (4):\penalty0 441
  -- 458, 2001.
\newblock URL
  \url{http://www.sciencedirect.com/science/article/pii/S0168927400000556}.

\bibitem[Stein(1999)]{b-Stein1999}
Michael~L. Stein.
\newblock \emph{{Interpolation of Spatial Data: Some Theory for Kriging}}.
\newblock Springer-Verlag New York, 1999.

\end{thebibliography}

\appendix
\section{The Mechanistic Hormonal Cycle Model}
\label{app:mech_model}

We provide here the equations for the mechanistic model of hormonal cycle from \citet{j-Clark2003}.

\begin{equation}
\resizebox{\textwidth}{!}
{$
        \begin{aligned}[c]
        \frac{dLH_{RP}}{dt} &= \frac{V_{0,LH}+\frac{ V_{1,LH} \cdot E_2(t)^8}{Km_{LH}^8+E_2(t)^8}}{1+\frac{P_4(t-\Delta_p)}{Ki_{LH,P}}}-\frac{k_{LH} \cdot \left[1+c_{LH,P} \cdot P_4(t)\right] \cdot LH_{RP}(t)}{1+c_{LH,E} \cdot E_2(t)} \\
        \frac{dLH}{dt} &=\frac{k_{LH}\cdot \left[1+c_{LH,P} \cdot P_4(t)\right]\cdot LH_{RP}(t)}{\nu \cdot \left[1+c_{LH,E} \cdot E_2(t)\right]} - a_{LH} \cdot LH(t)  \\
        \frac{dFSH_{RP}}{dt} &= \frac{V_{FSH}}{1+\frac{I_h(t-\Delta_{I_h})}{Ki_{FSH,I_h}}}-\frac{k_{FSH}\cdot\left[1+c_{FSH,P} \cdot P_4(t)\right] \cdot FSH_{RP}(t)}{1+ c_{FSH,E} \cdot E_2(t)^2} \\
        \frac{dFSH}{dt} &= \frac{k_{FSH}\cdot\left[1+c_{FSH,P} \cdot P_4(t)\right] \cdot FSH_{RP}(t)}{\nu \cdot \left[1+ c_{FSH,E} \cdot E_2(t)^2\right]} - a_{FSH} \cdot FSH(t) \\
        E_2(t)&=e_0+e_1 \cdot SeF(t) + e_2 \cdot PrF(t) + e_3 \cdot Lut_4(t) \\
        P_4(t)&=p_0+p_1 \cdot Lut_3 + p_2 \cdot Lut_4 \\
        I_h(t)&=h_0+h_1 \cdot PrF(t) + h_2 \cdot Lut_3 + h_3 \cdot Lut_4 \\
        \end{aligned}
        \qquad 
        \begin{aligned}[c]
        \frac{dRcF}{dt} &= b \cdot FSH(t)+\left[c_1 \cdot FSH(t) - c_2 \cdot LH(t)^{\alpha}\right] \cdot RcF(t) \\
        \frac{dSeF}{dt} &= c_2 \cdot LF(t)^{\alpha} \cdot RcF(t) + \left[c_3 \cdot LH(t)^{\beta}-c_4 \cdot LH(t) \right] \cdot SeF(t) \\
        \frac{dPrF}{dt} &= c_4 \cdot LH(t) \cdot SeF(t) - c_5 \cdot LH(t)^{\gamma} \cdot PrF(t) \\
        \frac{dSc1}{dt} &= c_5 \cdot LH(t)^{\gamma} \cdot PrF(t) - d_1 \cdot Sc1(t) \\
        \frac{dSc2}{dt} &= d_1 \cdot Sc1(t) - d_2 \cdot Sc2(t) \\
        \frac{dLut_1}{dt} &= d_2 \cdot Sc2(t) - k_1\cdot Lut_1(t) \\
        \frac{dLut_2}{dt} &= k_1 \cdot Lut_1(t) - k_2 \cdot Lut_2(t) \\
        \frac{dLut_3}{dt} &= k_2 \cdot Lut_2(t) - k_3 \cdot Lut_3(t) \\
        \frac{dLut_4}{dt} &= k_3 \cdot Lut_3(t) - k_4 \cdot Lut_4(t) \\
        \end{aligned}
        $}
\label{eq:hmc_model}
\end{equation}

\section{Gaussian Process Kernels}
\label{app:gpm_kernels}

For the work in this paper, we have focused on the combination of the rational quadratic (RQ) and periodic kernels (PE), \ie
\begin{equation}
k_{RQ}(\tau) = \left(1 + \frac{\tau^2}{2 \alpha l_{RQ}^2}\right)^{-\alpha} \qquad \text{ and } \qquad k_{PE}(\tau) = \exp\left(-2 \cdot \frac{\sin^2(\pi \tau \omega)}{l_{PE}^2}\right) \; .
\end{equation}
The rational quadratic kernel is a scale mixture of squared exponential kernels with different length-scales, while the standard periodic kernel can be explained by a mapping $u(x) = (\cos(x), \sin(x))$ through the squared exponential covariance function. We are interested in these two kernels for the female hormonal model, since we expect the data to be periodic and smooth (the RQ kernel is mean-squared differentiable and the most general representation for an isotropic kernel).

\section{Accuracy on predicting menstrual cycle phases}
\label{app:event_prediction_tables}

Tables \ref{tab:event_prediction_rate} and \ref{tab:event_prediction_noise} show the accuracy of predicting menstrual cycle phases using GPMs. Provided results correspond to training intervals where exactly two peaks of the cycle are included; longer training intervals produce similar results. Table \ref{tab:event_prediction_rate} illustrates the impact of downsampling, while Table \ref{tab:event_prediction_noise} shows the impact of noise.

Accuracy is shown as the the ratio of days predicted for hormonal events (peak or valley) to ground truth events. Two sampling strategies are considered: \texttt{agnostic} sampling, where data is acquired at regular intervals without any knowledge about the underlying hormonal cycle, and \texttt{informed} sampling, where at least one sample from the peak (or valley) of the hormonal cycle is included.

\begin{table}[!h]
	\begin{center}
	\resizebox{0.99\textwidth}{!}{
		\begin{tabular}{*{7}{|c}|}
	\hline
	Sampling period \cellcolor[gray]{0.6} & Sampling type \cellcolor[gray]{0.6} & $LH$ peak precision \cellcolor[gray]{0.6} & $FSH$ peak precision \cellcolor[gray]{0.6} & $E_2$ valley precision \cellcolor[gray]{0.6} & $P_4$ peak precision \cellcolor[gray]{0.6} & $I_h$ peak precision \cellcolor[gray]{0.6} \\ \hline 
	1 day & Agnostic & 2.0/2.0 & 2.0/2.0 & 5.0/5.0 & 7.0/7.0 & 7.0/7.0 \\ \hline 
	1 day \cellcolor[gray]{0.9}& Informed \cellcolor[gray]{0.9} & 2.0/2.0 \cellcolor[gray]{0.9}& 2.0/2.0 \cellcolor[gray]{0.9}& 5.0/5.0 \cellcolor[gray]{0.9}& 7.0/7.0 \cellcolor[gray]{0.9}& 7.0/7.0 \cellcolor[gray]{0.9} \\ \hline 
	2 days & Agnostic & 1.74/2.0 & 1.74/2.0 & 4.47/5.0 & 6.79/7.0 & 5.95/7.0 \\ \hline 
	2 days \cellcolor[gray]{0.9}& Informed \cellcolor[gray]{0.9}& 1.32/2.0 \cellcolor[gray]{0.9}& 1.37/2.0 \cellcolor[gray]{0.9}& 3.75/5.0 \cellcolor[gray]{0.9}& 6.74/7.0 \cellcolor[gray]{0.9}& 6.06/7.0 \cellcolor[gray]{0.9}\\ \hline 
	4 days & Agnostic & 0.04/2.0 & 0.04/2.0 & 3.84/5.0 & 5.88/7.0 & 5.29/7.0 \\ \hline 
	4 days \cellcolor[gray]{0.9}& Informed \cellcolor[gray]{0.9}& 0.32/2.0 \cellcolor[gray]{0.9}& 0.08/2.0 \cellcolor[gray]{0.9}& 2.36/5.0 \cellcolor[gray]{0.9}& 6.68/7.0 \cellcolor[gray]{0.9}& 6.24/7.0 \cellcolor[gray]{0.9}\\ \hline 
	6 days & Agnostic & 0.56/2.0 & 0.48/2.0 & 2.32/5.0 & 4.24/7.0 & 4.79/7.0 \\ \hline 
	6 days \cellcolor[gray]{0.9}& Informed \cellcolor[gray]{0.9}& 0.64/2.0 \cellcolor[gray]{0.9}& 0.44/2.0 \cellcolor[gray]{0.9}& 3.16/5.0 \cellcolor[gray]{0.9}& 6.36/7.0 \cellcolor[gray]{0.9}& 6.16/7.0 \cellcolor[gray]{0.9}\\ \hline 
	\end{tabular}
	}
	\vspace*{0.5cm}
	\caption{GP prediction of hormonal phases of interest, with no observation noise.}
	\label{tab:event_prediction_rate}
	\end{center}
\end{table}

\begin{table}[!h]
	\begin{center}
		\resizebox{0.99\textwidth}{!}{
			\begin{tabular}{*{8}{|c}|}
				\hline
				Noise ratio \cellcolor[gray]{0.6} & Sampling period \cellcolor[gray]{0.6} & Sampling type \cellcolor[gray]{0.6} & $LH$ peak precision \cellcolor[gray]{0.6} & $FSH$ peak precision \cellcolor[gray]{0.6} & $E_2$ valley precision \cellcolor[gray]{0.6} & $P_4$ peak precision \cellcolor[gray]{0.6} & $I_h$ peak precision \cellcolor[gray]{0.6} \\ \hline 
				$\frac{\sigma_{\epsilon}}{\sigma_{f}}=0.0$ & 1 day & Agnostic & 2.0/2.0 & 2.0/2.0 & 5.0/5.0 & 7.0/7.0 & 7.0/7.0 \\ \hline 
				$\frac{\sigma_{\epsilon}}{\sigma_{f}}=0.0$ & 1 day\cellcolor[gray]{0.9}& Informed \cellcolor[gray]{0.9} & 2.0/2.0 \cellcolor[gray]{0.9}& 2.0/2.0 \cellcolor[gray]{0.9}& 5.0/5.0 \cellcolor[gray]{0.9}& 7.0/7.0 \cellcolor[gray]{0.9}& 7.0/7.0 \cellcolor[gray]{0.9} \\ \hline 
				$\frac{\sigma_{\epsilon}}{\sigma_{f}}=0.0$ & 2 days & Agnostic & 1.74/2.0 & 1.74/2.0 & 4.47/5.0 & 6.79/7.0 & 5.95/7.0 \\ \hline 
				$\frac{\sigma_{\epsilon}}{\sigma_{f}}=0.0$ & 2 days \cellcolor[gray]{0.9}& Informed \cellcolor[gray]{0.9}& 1.32/2.0 \cellcolor[gray]{0.9}& 1.37/2.0 \cellcolor[gray]{0.9}& 3.37/5.0 \cellcolor[gray]{0.9}& 6.73/7.0 \cellcolor[gray]{0.9}& 6.06/7.0 \cellcolor[gray]{0.9}\\ \hline 
				
				$\frac{\sigma_{\epsilon}}{\sigma_{f}}=0.01$ & 1 day & Agnostic & 2.0/2.0 & 2.0/2.0 & 4.5/5.0 & 7.0/7.0 & 7.0/7.0\\ \hline 
				$\frac{\sigma_{\epsilon}}{\sigma_{f}}=0.01$ & 1 day \cellcolor[gray]{0.9}& Informed \cellcolor[gray]{0.9} & 2.0/2.0 \cellcolor[gray]{0.9}& 2.0/2.0 \cellcolor[gray]{0.9}& 4.66/5.0 \cellcolor[gray]{0.9}& 7.0/7.0 \cellcolor[gray]{0.9}& 4.67/7.0 \cellcolor[gray]{0.9}\\ \hline  
				$\frac{\sigma_{\epsilon}}{\sigma_{f}}=0.01$ & 2 days & Agnostic & 1.53/2.0 & 1.63/2.0 & 4.63/5.0 & 5.89/7.0 & 4.36/7.0\\ \hline 
				$\frac{\sigma_{\epsilon}}{\sigma_{f}}=0.01$ & 2 days \cellcolor[gray]{0.9}& Informed \cellcolor[gray]{0.9}& 1.13/2.0 \cellcolor[gray]{0.9}& 1.44/2.0 \cellcolor[gray]{0.9}& 4.18/5.0 \cellcolor[gray]{0.9}& 6.5/7.0 \cellcolor[gray]{0.9}& 5.68/7.0 \cellcolor[gray]{0.9}\\ \hline 
				
				$\frac{\sigma_{\epsilon}}{\sigma_{f}}=0.1$ & 1 day & Agnostic & 1.0/2.0 & 1.5/2.0 & 1.75/5.0 & 6.5/7.0 & 7.0/7.0 \\ \hline 
				$\frac{\sigma_{\epsilon}}{\sigma_{f}}=0.1$ & 1 day \cellcolor[gray]{0.9}& Informed \cellcolor[gray]{0.9} & 2.0/2.0 \cellcolor[gray]{0.9}& 1.0/2.0 \cellcolor[gray]{0.9}& 1.67/5.0 \cellcolor[gray]{0.9}& 6.67/7.0 \cellcolor[gray]{0.9}& 6.0/7.0 \cellcolor[gray]{0.9} \\ \hline 
				$\frac{\sigma_{\epsilon}}{\sigma_{f}}=0.1$ & 2 days & Agnostic & 1.74/2.0 & 1.68/2.0 & 3.63/5.0 & 6.05/7.0 & 6.01/7.0 \\ \hline 
				$\frac{\sigma_{\epsilon}}{\sigma_{f}}=0.1$ & 2 days \cellcolor[gray]{0.9}& Informed \cellcolor[gray]{0.9}& 1.13/2.0 \cellcolor[gray]{0.9}& 1.56/2.0 \cellcolor[gray]{0.9}& 3.0/5.0 \cellcolor[gray]{0.9}& 5.51/7.0 \cellcolor[gray]{0.9}& 5.5/7.0 \cellcolor[gray]{0.9}\\ \hline 
			\end{tabular}
		}
		\vspace*{0.5cm}
		\caption{GP prediction of hormonal phases of interest, for different observation noise levels.}
		\label{tab:event_prediction_noise}
	\end{center}
\end{table}

\end{document}